\documentclass[11pt, a4paper, logo, copyright, nonumbering]{kwaipilot}
\usepackage{dblfloatfix}
\usepackage{ulem}
\usepackage{caption}
\usepackage{dramatist}
\usepackage{xspace}
\usepackage{pifont} %
\usepackage{multirow}
\usepackage{tcolorbox}
\usepackage{xltabular}
\usepackage{longtable}
\usepackage{hyperref}
\usepackage{booktabs}
\usepackage{array}
\usepackage{xcolor}
\definecolor{lightgray}{RGB}{110,110,110}
\usepackage{hyperref}
\usepackage{footnote}
\usepackage{geometry}
\usepackage{tcolorbox}
\usepackage{caption}
\usepackage{rotating}
\usepackage{graphicx}
\usepackage{subcaption}
\usepackage{float}
\usepackage{amsfonts}
\usepackage{amsmath}
\usepackage{amssymb}
\usepackage{lineno}
\usepackage{multirow}
\usepackage{adjustbox}
\usepackage{setspace} 

\usepackage[bottom]{footmisc}

\usepackage{CJKutf8}
\usepackage{setspace}

\usepackage{dsfont}
\usepackage{array} %
\usepackage{tabularx} %
\usepackage{xcolor} %
\usepackage{tabularx}
\usepackage{booktabs}

\usepackage{lipsum}  %
\usepackage{multicol} %

\usepackage{indentfirst} 
\usepackage[sort&compress,numbers]{natbib}

\usepackage{makecell}

\usepackage{titlesec}
\titlespacing*{\paragraph}{0pt}{0.3ex plus 0.2ex minus .2ex}{1em}

\makeatletter
\def\@BTrule[#1]{%
  \ifx\longtable\undefined
    \let\@BTswitch\@BTnormal
  \else\ifx\hline\LT@hline
    \nobreak
    \let\@BTswitch\@BLTrule
  \else
     \let\@BTswitch\@BTnormal
  \fi\fi
  \global\@thisrulewidth=#1\relax
  \ifnum\@thisruleclass=\tw@\vskip\@aboverulesep\else
  \ifnum\@lastruleclass=\z@\vskip\@aboverulesep\else
  \ifnum\@lastruleclass=\@ne\vskip\doublerulesep\fi\fi\fi
  \@BTswitch}
\makeatother

\addto\extrasenglish{
}

 {\begin{list}{}%
         {\setlength{\leftmargin}{#1}}%
         \item[]%
 }
 {\end{list}}
 
\reportnumber{001} %

\title{\centering KAT-Coder Technical Report }

\author{
Zizheng Zhan$^{*\,\dag}$, Ken Deng$^{*\,\dag}$, Jinghui Wang$^{*}$, Xiaojiang Zhang$^{*}$, Huaixi Tang$^{*}$, Minglei Zhang$^{*}$, Zhiyi Lai$^{*}$, Haoyang Huang$^{*}$, Wen Xiang$^{*}$, Kun Wu$^{*}$, Wenhao Zhuang, Shaojie Wang, Shangpeng Yan, Kepeng Lei, Zongxian Feng, Huiming Wang, Zheng Lin, Mengtong Li, Mengfei Xie, Yinghan Cui, Xuxing Chen, Chao Wang, Weihao Li, Wenqiang Zhu, Jiarong Zhang, Jingxuan Xu, Songwei Yu, Yifan Yao, Xinping Lei, C. Zhang, Han Li, Junqi Xiong, Zuchen Gao, Dailin Li, Haimo Li, Jiaheng Liu, Yuqun Zhang, Junyi Peng, Haotian Zhang$^{\dag}$, Bin Chen \\
    Kwaipilot Team \\
    \texttt{\{zhanzizheng, dengken, zhanghaotian\}@kuaishou.com} \\
}

\begingroup
\footnotetext{
$^*$Equal contribution. $^\dag$Corresponding author. }
\endgroup

\renewcommand{\arraystretch}{1.2} 

\begin{abstract}
Recent advances in large language models (LLMs) have enabled rapid progress in agentic coding, where models autonomously reason, plan, and act within interactive software development workflows. Yet, bridging the gap between static text-based training and dynamic real-world execution remains a core challenge.
In this technical report, we present KAT-Coder, a large-scale agentic code model trained through a four-stage curriculum: Mid-Term Training, Supervised Fine-Tuning (SFT), Reinforcement Fine-Tuning (RFT), and Agentic Reinforcement Learning (Agentic RL).
The Mid-Term stage enriches reasoning, planning, and reflection abilities via a corpus of real software-engineering data and synthetic agentic trajectories. The SFT stage constructs a million-sample dataset balanced across 20+ programming languages, 10 development contexts, and 10 task archetypes, enabling cross-domain generalization. The RFT stage introduces a multi-ground-truth reward formulation for stable and sample-efficient policy optimization.
Building upon this, the final Agentic RL phase integrates Trie-Packed Training with Difficulty- and Entropy-Aware Advantage Rescaling, achieving efficient multi-trajectory optimization, enhanced exploration, and robust policy diversity.
Together, these stages establish a unified framework that bridges reasoning enrichment, structured supervision, reinforcement alignment, and deployment robustness. KAT-Coder attains strong and balanced results across reasoning, coding, and agentic benchmarks, forming a deployable foundation for real-world intelligent coding agents. 
Our \textbf{KAT} series 32B model, \textbf{KAT-Dev}, has been open-sourced on \includegraphics[height=1.0em]{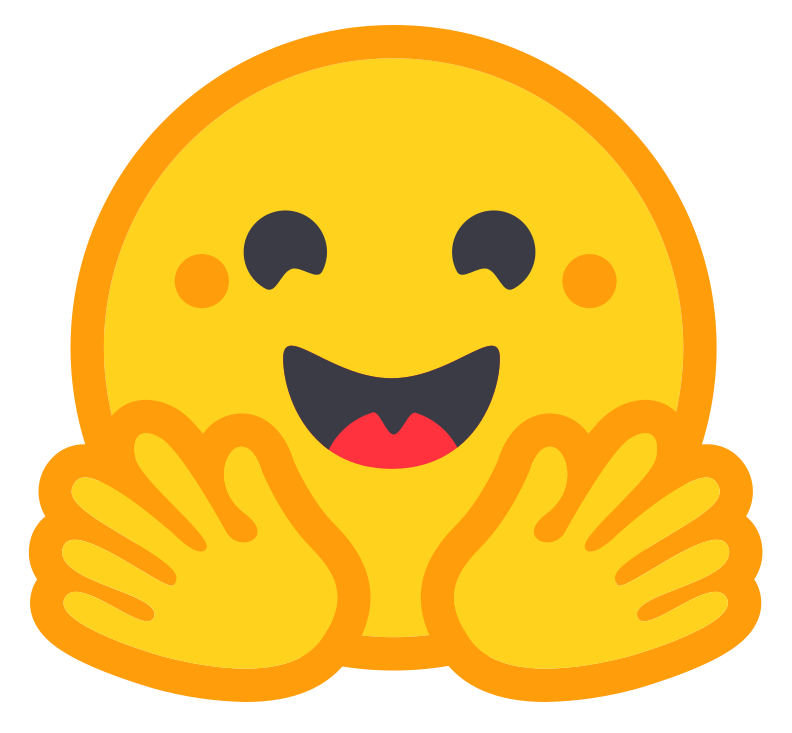} \url{https://huggingface.co/Kwaipilot/KAT-Dev}.
\end{abstract}

\begin{document}

\maketitle

\newpage
\section{Introduction}

Recent advances in Large Language Models (LLMs) have catalyzed a shift from static text generation toward agentic intelligence, where models autonomously reason, plan, and act within dynamic environments. In software engineering, this transformation manifests as agentic coding—a paradigm in which models evolve from passive code generators into collaborative problem solvers. Despite remarkable progress, a core challenge persists: bridging the gap between static, text-based training and interactive, real-world execution. Conventional code models, typically trained on massive but inert text corpora, lack the adaptive reasoning and contextual control required to operate reliably in live integrated development environments (IDEs)~\cite{sweagentagent,cursor,codex,roocode,cline,anthropicclaude,wang2025openhands,codeflicker2025}.

Early frameworks such as Codex~\cite{codex}, CodeLlama~\cite{roziere2023codellama}, and DeepSeekCoder~\cite{deepseekcoder2024} established the foundation for code generation, yet they remain limited to single-turn, instruction-following behavior. More recent efforts, including SWE-Agent~\cite{sweagentagent}, OpenHands~\cite{wang2025openhands}, and Claude Code~\cite{anthropicclaude}, introduce planning and tool-use capabilities, signaling a broader trend toward agentic execution. However, these systems are often constrained by narrow domain coverage, short reasoning horizons, and homogeneous datasets that fail to capture the diversity of real software engineering workflows. Consequently, their performance deteriorates when transferred from benchmark environments to production-grade systems characterized by heterogeneous tools, long-term dependencies, and frequent context shifts.

To address these limitations, we introduce \textbf{KAT-Coder}—a large-scale agentic code model designed to unify reasoning, planning, reinforcement, and deployment robustness within a single training framework. Specifically, KAT-Coder is developed through a four-stage hierarchical curriculum that progressively enhances the model’s cognitive and operational competence:

\begin{itemize}
    \item \textbf{Mid-Term Training} — Broadens reasoning, planning, and reflective thinking abilities through a combination of real software engineering corpora and synthetic agentic trajectories, forming a bridge between general pretraining and code-oriented supervision.
    \item \textbf{Supervised Fine-Tuning (SFT)} — Constructs a million-sample dataset spanning over 20 programming languages, 10 development contexts, and 10 task archetypes, ensuring balanced coverage and strong cross-domain generalization.
    \item \textbf{Reinforcement Fine-Tuning (RFT)} — Introduces a multi-ground-truth reward formulation and relative evaluation scheme for stable and sample-efficient policy optimization.
    \item \textbf{Agentic Reinforcement Learning (Agentic RL)} — Implements Trie-Packed Training for high-throughput multi-trajectory optimization and Difficulty- and Entropy-Aware Advantage Rescaling to maintain exploration diversity, prevent entropy collapse, and enhance policy robustness.
\end{itemize}

This curriculum reflects a closed-loop design philosophy: cognitive enrichment precedes structured supervision, which in turn grounds reinforcement alignment and culminates in real-world adaptation. Through this progressive alignment, KAT-Coder evolves from a general LLM into a deployable agentic developer capable of long-context reasoning, dynamic tool management, and collaborative problem solving across complex software workflows.

\begin{figure}
    \centering
    \includegraphics[width=1\linewidth]{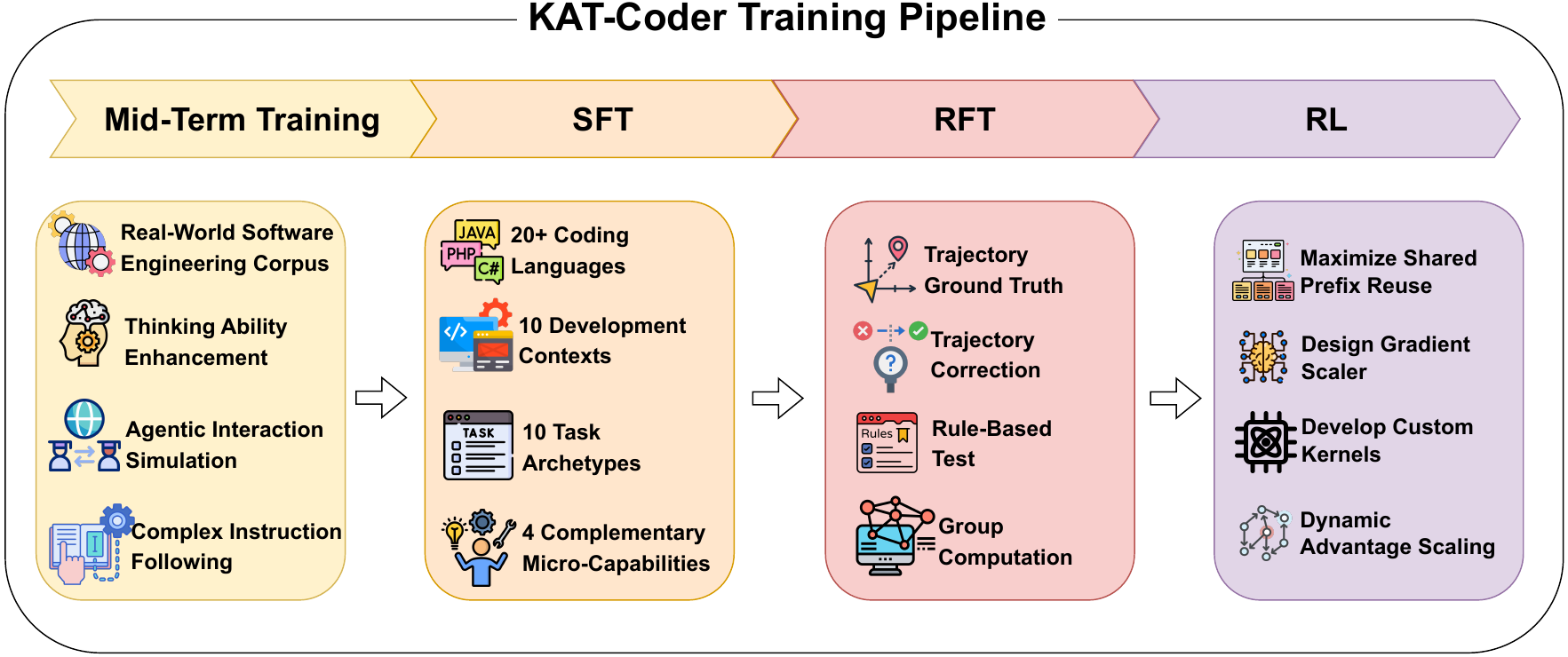}
    \caption{KAT-Coder Training Pipeline}
    \label{fig:Pipline}
\end{figure}

\section{Mid-Term Training}

The Agentic capability of a model represents a composite form of intelligence that integrates multiple dimensions—tool use, instruction following, long-context reasoning, code generation, and multi-turn dialogue. These dimensions collectively determine the model’s capacity for autonomous decision-making and adaptive interaction within real-world coding environments. To fully unlock these abilities before introducing real Agentic supervision data, we perform an extensive Mid-Term Training phase designed to broaden the model’s reasoning, planning, and interactive scope, thereby establishing a solid foundation for subsequent Code-Oriented SFT.

\paragraph{Training Recipe Overview}
Our Mid-Term Training recipe spans a diverse set of domains and task formulations, targeting both the structural and cognitive aspects of Agentic behavior. The design consists of four main components:

(1) \textbf{Real-World Software Engineering Corpus:} We collected approximately 20B tokens of real user programming data from GitHub, including pull requests, issues, commits, and corresponding code diff patches. This corpus captures authentic human–code interactions and evolution patterns across collaborative development workflows\cite{SWESwiss2025,muennighoff2023octopack,badertdinov2025swerebenchautomatedpipelinetask,pan2025trainingsoftwareengineeringagents,yang2025swesmithscalingdatasoftware,zhang2025swebenchgoeslive,guertler2025textarena,korgym}.
(2) \textbf{Reasoning and Reflection Enhancement:} To activate the model’s reasoning, planning, and reflective thinking capabilities, we utilized advanced open-source reasoning models to generate chain-of-thought trajectories for solving complex software engineering problems, competition-level STEM challenges, and logical puzzles. This promotes robust multi-domain reasoning and systematic thinking\cite{tian2025correctanswersequaldistillation,2025iithought,chen2025acereasonnemotronadvancingmathcode,he2025skyworkopenreasoner1,deng2025hipo}.
(3) \textbf{Agentic Interaction Simulation:} To familiarize the model with real Agentic interaction patterns, we constructed simulated environments that synthesize trajectories reflecting Plan–Action–Observation loops under diverse contexts. These trajectories train the model to dynamically adapt its plans and decisions based on environmental feedback\cite{sweagentagent,cursor,codex,roocode,cline,anthropicclaude,wang2025openhands}.
(4) \textbf{Complex Instruction Following and Constraint Alignment:} We further curated instruction datasets with verifiable logical and structural constraints, enabling the model to improve its consistency, controllability, and robustness in handling complex, multi-condition instructions\cite{yang2025ifevalcodecontrolledcodegeneration,zhang2025inverseifevalllmsunlearn,ouyang2022traininglanguagemodelsfollow,shi2024instructiontuninglossinstructions,wang2025codeifbenchevaluatinginstructionfollowingcapabilities}.

This Mid-Term Training recipe substantially strengthens the model’s foundational reasoning, reflection, and interaction capabilities, providing a crucial bridge between general pretraining and Agentic SFT. It serves as a decisive stage for expanding the upper limit of the model’s cognitive and operational competence in real-world code-oriented tasks.

Existing open-source corpora for agentic coding tasks exhibit a strong distributional bias, with the majority of samples focusing on Python-based bug-fixing activities. However, real-world programming practices extend far beyond this narrow scope, encompassing a wide range of languages, development contexts, and task archetypes. To enable robust generalization and realistic adaptation in practical engineering scenarios, we systematically redesign the dataset along three orthogonal axes—programming languages, development contexts, and task types—ensuring balanced coverage and diversity across them.

\paragraph{Data Sources and Statistical Analysis}

Our dataset construction is grounded in large-scale mining and analysis of open-source repositories and community discussions from GitHub and Stack Overflow. By examining commit histories, code diffs, review comments, and Q\&A threads, we extracted and summarized patterns of user activity and developer intent. This statistical foundation enables principled sampling and categorization across language, context, and task dimensions.

\paragraph{Programming Language Dimension}

To reflect the diversity of modern software ecosystems, we cover over twenty mainstream programming languages. Beyond high-frequency languages such as Python, Java, TypeScript, JavaScript, C, C++, C\#, Kotlin, Go, Rust, PHP, and Ruby, we extend coverage to include Swift, Objective-C, Scala, R, Shell/Bash, SQL, MATLAB, Dart, Lua, Elixir, Haskell, and Perl. This broad spectrum—from scripting to systems programming—ensures generalization across heterogeneous language paradigms.

\paragraph{Development Context Dimension}

We identify ten representative development contexts through statistical analysis of real-world coding activities, encompassing the full spectrum of software engineering practices: application development, system and infrastructure development, UI/UX engineering, data science and engineering, database systems, machine learning and artificial intelligence, algorithm design and analysis, testing and debugging, system architecture and maintenance, and specialized programming domains.
Balanced sampling across these contexts prevents domain overrepresentation and enhances the dataset’s robustness for cross-context generalization.

\paragraph{Task Type Dimension}

At the task level, we distill ten fundamental archetypes that capture the essential forms of software development behavior: implementation, modification and feature enhancement, debugging and bug fixing, refactoring, performance optimization, code explanation and documentation, code analysis, code generation, test case generation, and configuration and deployment.

This taxonomy spans the full development lifecycle—from problem formulation to solution deployment—capturing both cognitive and operational aspects of real-world programming.

\paragraph{Dataset Scale and Distribution}

The resulting SFT corpus comprises over one million samples, covering a rich combination of languages, contexts, and task types. Such diversity ensures balanced representation of programming practices and provides a solid supervised foundation for subsequent reinforcement fine-tuning (RFT) and reinforcement learning (RL) stages.

\subsection{Reinforcement-to-Deployment Adaptation: Bridging Research Agents and Real-World Workflows}

\paragraph{Motivation}

Existing datasets for code agents in the supervised fine-tuning (SFT) stage are primarily derived from research-oriented frameworks such as SWE-Agent\cite{sweagentagent}, which rely on linear, single-session dialogues and homogeneous operation pipelines. While these datasets are effective for controlled academic evaluation, they fail to capture the complexity of real-world agentic environments.

In practical software engineering, agents must operate within heterogeneous toolchains, dynamically manage long-horizon dependencies, and adapt to non-linear conversational trajectories involving frequent context switches and multi-turn reasoning. This gap between research benchmarks and production-grade workflows leads to a significant distribution mismatch, limiting the generalization ability of agentic code models when deployed in real development systems.

\paragraph{Data Construction across Production Environments}

To bridge this research–deployment gap, we construct a new generation of Agentic Workflow training data by integrating our early-stage KAT-Coder models with production-grade IDE-based systems such as Claude Code\cite{anthropicclaude}, Cline\cite{cline}, Roo Code\cite{roocode}, and CodeFlicker\cite{codeflicker2025}. These environments offer realistic execution traces, tool invocations, and iterative human–agent interactions, enabling the construction of data that reflects true software development dynamics.

This integration yields trajectories that are far more diverse and realistic but also introduces new training challenges.

\paragraph{Training Challenges in Production-Grade Trajectories}

Production workflows differ substantially from benchmark settings in two major aspects: (1) Expanded Tool Spectrum — Real-world agents interact with dozens of heterogeneous tools (e.g., debuggers, linters, package managers), leading to frequent erroneous or redundant tool calls. (2) Non-Linear Context Boundaries — Compression checkpoints, context truncation, and mode switches (e.g., between coding, planning, and execution) introduce branching points that disrupt the continuity of dependency chains.
These challenges make direct imitation learning unstable, as gradient propagation from noisy tool calls or broken trajectories can degrade convergence and overfit to spurious behaviors.

\paragraph{Methodology: Error-Masked SFT and Tree-Structured Trajectory Training}

To address these issues, we adopt two complementary strategies designed for agentic fine-tuning under complex tool and context dynamics: (1) Error-Masked SFT (EM-SFT) We leverage execution feedback logs to identify tool-use failures and selectively mask gradients from erroneous tool calls. This prevents error amplification during backpropagation while retaining the model’s exposure to self-corrective reasoning signals. (2) Tree-Structured Trajectory Training (TST) We decompose multi-branch trajectories into locally coherent subtrees defined by context compression boundaries and mode transitions. Within each subtree, standard supervised fine-tuning is performed independently, ensuring stable optimization and improved temporal consistency.
Together, these strategies enable the policy model to learn from realistic, production-grade trajectories without sacrificing training stability or semantic coherence. This forms the foundation for aligning the model’s behavior with human engineering workflows—an essential step toward fully deployable agentic coding systems.

\section{RFT}

During the reinforcement learning (RL) phase, we introduce multiple trajectory ground truths as reference signals to enhance rollout efficiency and training stability. Traditional RL approaches typically rely on an absolute reward computed directly from the model’s output, which is highly sensitive to the reward scale and often leads to unstable optimization or inefficient sample utilization. 

To address this, we propose a relative evaluation framework, where the model is optimized based on the discrepancy between generated samples and ground-truth trajectories rather than absolute reward magnitudes. This transformation stabilizes the training process and significantly improves sampling efficiency.

\begin{figure}
    \centering
    \includegraphics[width=1\linewidth]{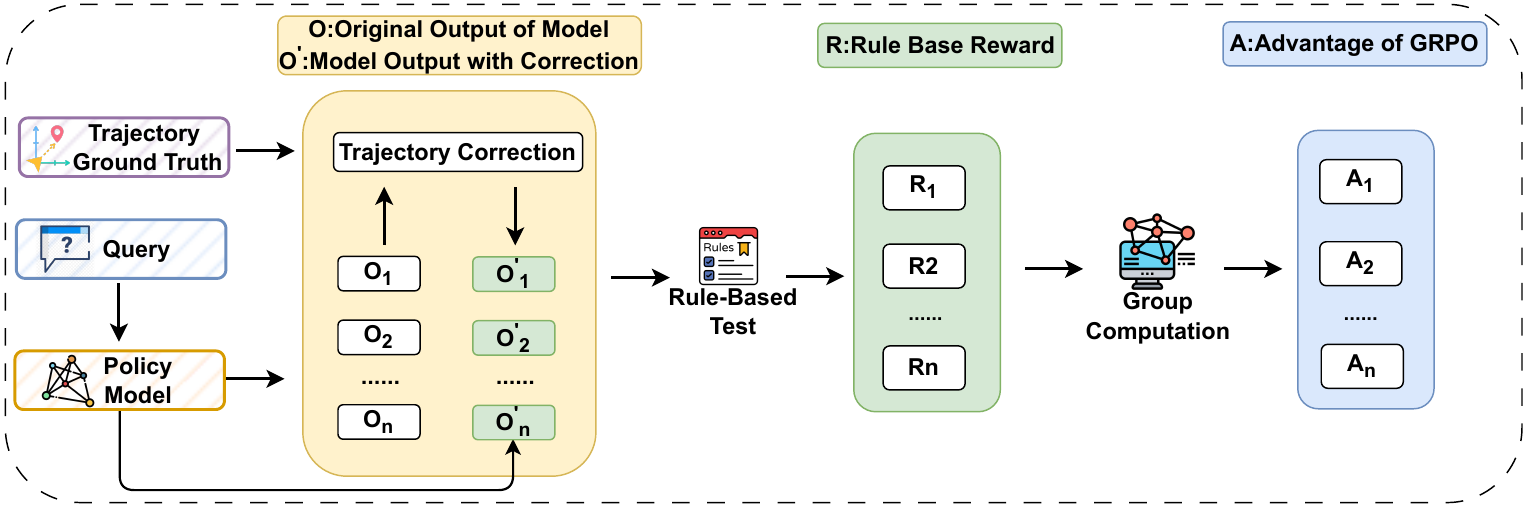}
    \caption{Method overview of RFT. (1)Trajectory Ground Truth: provides evaluation and correction signals for generated trajectories. (2) Trajectory Correction: performs online correction for outputs deviating from reference trajectories. (3) Rule-Based Test: generates stable and interpretable rule-based reward signals. (4) Group Computation: normalizes and aggregates group-level advantages within the GRPO framework.}
    \label{fig:RFT}
\end{figure}

As illustrated in Figure \ref{fig:RFT}, given an input query \( q \), the policy model \( \pi_\theta \) produces a sequence of trajectory outputs \( o_1, o_2, \dots, o_n \). A set of trajectory ground truths is then used for \textbf{Trajectory Correction}, generating corrected outputs \( o'_i \). These are evaluated through a \textbf{Rule Base Test}, which yields rule-based rewards \( r_i \). Subsequently, the \textbf{Group Computation} stage aggregates and normalizes the sample groups under the GRPO \cite{shao2024deepseekmath} framework to produce final advantages \( A_i \) for policy gradient updates. This hierarchical correction and grouping mechanism enables the policy to converge towards semantically correct and structurally consistent generation behaviors.

\paragraph{Training Stability and Sample Efficiency}
By replacing absolute reward computation with relative discrepancy estimation, the method effectively mitigates instability caused by model fluctuations or reward scale drift. Additionally, early termination and resampling mechanisms for trajectories that deviate significantly from ground truths improve overall sample utilization and rollout efficiency.

\paragraph{Empirical Insights}
This design yields substantial improvements during RL training, including:
(1) More stable reward signals, reducing scale drift and enhancing convergence.
(2) Higher sample efficiency, by filtering invalid rollouts through correction and resampling.
(3) Better semantic alignment, as generated trajectories align more closely with human-validated ground truths.

\section{Agentic RL}

\subsection{Trie Packed Training}

\begin{figure}
    \centering
    \includegraphics[width=1\linewidth]{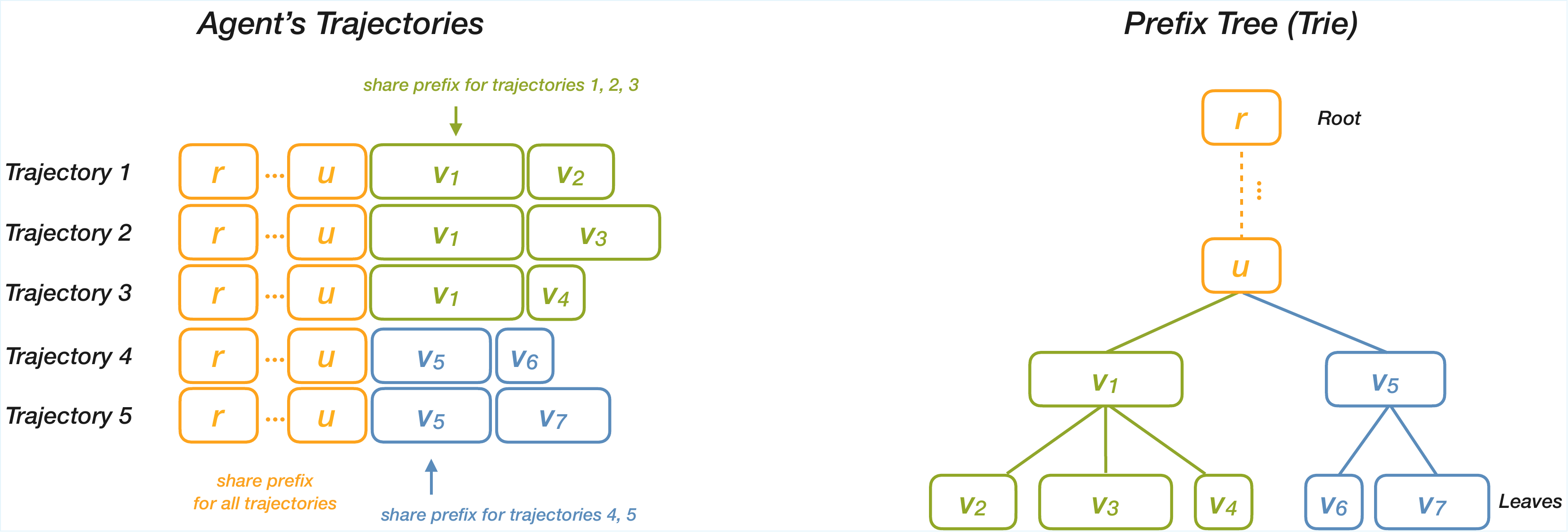}
    \caption{Overview of Trie-Packed Training in agentic RL.}
    \label{fig:RL}
\end{figure}

In agentic LLM training, an agent's behavior is often highly diverse, so its trajectory history is usually not a linear trajectory like prompt1 → response1 → prompt2 → response2. Instead, a single task can produce multiple trajectories, many of which share common prefixes. As shown in figure \ref{fig:RL}, these trajectories can be naturally organized into a prefix tree (Trie).

In this scenario, we realized that the computation of prefixes shared across multiple trajectories could be performed just once, which would significantly improve training throughput. This idea is similar to the prefix caching used for LLM inference. However, due to the involvement of back propagation during training, we cannot directly reuse cached results; doing so would ignore the gradient contributions from suffix tokens to the prefix tokens, leading to incorrect computations.
Our goal is computing each shared prefix only once during both forward and backward and thus improve agentic LLM training throughput.
To achieve this, we proposed Trie Packed Training paradigm\cite{wang2025trie}:

(1) Maximize shared prefix reuse through Trie Packing: Since there are many training trajectories, even after merging them into a tree structure, we usually can’t fit the entire tree into one batch due to GPU memory constraints. Using a combination of dynamic programming and greedy algorithms, we proposed a practical approach to pack the tree under memory constraints which maximizes the reuse of shared prefixes.

(2) Design Gradient Scaler to ensure correct gradient computation: During back propagation, the gradient contribution of a shared prefix differs across different trajectories. We implemented a tree-structured gradient scaler that mathematically ensures each shared prefix contributes correctly to the gradients.

(3) Develop custom kernels: We implemented an efficient shared-prefix mask attention and modified position embedding for flattened trie pack data, ensuring both training correctness and high throughput.
\subsection{Enhancing Exploration via Difficulty- and Entropy-Aware Advantage Rescaling}

In policy gradient reinforcement learning, the advantage function $A(s,a)$ determines each sample’s contribution to the update, which can be computed as:
\[
\theta \leftarrow \theta + \eta \, \nabla_\theta \log \pi_\theta(a|s) \, A(s,a)
\]

Larger advantages increase a sample’s influence, but standard GRPO tends to assign the largest advantages to medium-difficulty tasks, limiting exploration on very easy or hard tasks and potentially causing entropy collapse. To address this, we propose a difficulty- and entropy-aware advantage scaling method that adjusts group-level advantages according to task difficulty. For the $i$-th task group, difficulty is defined based on the average success rate:
\[
D_i = 1 - r_i
\]
where $r_i$ denotes the group’s average success rate. Higher $D_i$ indicates tasks that are not yet mastered and require stronger optimization, while lower $D_i$ corresponds to easier tasks with reduced emphasis. The group-level scaling factor is defined as:
\[
\alpha_i^{\text{group}} = 1 + \lambda (D_i - \bar{D})
\]
where $\bar{D}$ is the average difficulty of the current batch and $\lambda>0$ controls the magnitude of scaling.

Within each group, we further adjust the advantage based on the policy entropy of each sample. High-entropy samples indicate greater uncertainty in the model’s action selection and thus higher exploratory value, whereas low-entropy samples correspond to more certain decisions. The sample-level scaling factor is defined as:
\[
\beta_{ij}^{\text{sample}} = 1 + \mu (H_{ij} - \bar{H}_i)
\]
where $H_{ij}$ is the policy entropy of sample $j$ in group $i$, $\bar{H}_i$ is the group’s average entropy, and $\mu>0$ controls the scaling strength.

Finally, the difficulty- and entropy-aware scaled advantage for each sample is given by:
\[
A'_{ij} = \alpha_i^{\text{group}} \, \beta_{ij}^{\text{sample}} \, A_{ij}
\]

This method dynamically allocates training resources to amplify high-difficulty tasks and emphasize high-entropy samples, maintaining policy diversity, preventing entropy collapse, and enhancing exploration, robustness, and generalization.

\section{Model Evaluation and Comparative Analysis}

We comprehensively evaluate KAT-Coder across diverse benchmarks covering instruction following (IFEval\cite{ifeval}), tool invocation (TAU2-Bench Retail\cite{tau2bench}), mathematical reasoning (AIME 2025), code generation (LiveCodeBench V6\cite{livecodebench}, HumanEval\cite{humaneavl}), general knowledge (GPQA-Diamond\cite{gpqa}), and agentic coding (SWE-Bench-Verified\cite{jimenez2024swebenchlanguagemodelsresolve}). The results, summarized in Table \ref{tab:katcoder-benchmarks}, are compared with leading contemporary large models including Qwen3-Coder-480B\cite{yang2025qwen3,qwencoder}, Kimi-k2-0905\cite{kimik2}, and Claude 4 Sonnet\cite{claude4}.


\begin{table}[h]
  \centering
  \begingroup
  \setlength{\tabcolsep}{4pt} 
  \renewcommand{\arraystretch}{0.9} 
  \begin{tabular}{lcccc}
    \toprule
    \textbf{Benchmark} & \textbf{KAT-Coder} & \textbf{Qwen3-Coder-480B} & \textbf{Kimi-k2-0905} & \textbf{Claude 4 Sonnet} \\
    \midrule
    IFEval                    & 86.0 & 84.8 & \textbf{89.3} & 88.2 \\
    TAU2-Bench Retail         & 62.3 & 57.9  & 56.5  & \textbf{64.2} \\
    AIME 2025                & \textbf{72.5}  & 44.3 & 49.5 & 70.5  \\
    LiveCodeBench V6         & 48.2 & 48.2 & \textbf{48.9} & 46.5 \\
    HumanEval               & 96.3 & 95.1 & 88.4 & \textbf{98.2} \\
    GPQA-Diamond            & 68.2 & 60.6 & \textbf{70.2}  & 68.7 \\
    SWE-Bench-Verified      & \textbf{73.4}  & 69.6  & 65.8  & 72.7  \\
    \bottomrule
  \end{tabular}
  \endgroup
  \caption{Benchmark results comparing KAT-Coder with contemporary large models.}
  \label{tab:katcoder-benchmarks}
\end{table}

Across a broad range of evaluation tracks, KAT-Coder demonstrates consistently strong and balanced performance. It achieves competitive results in instruction following, tool invocation, mathematical reasoning, code generation, and general knowledge, matching or surpassing leading proprietary and open-source baselines such as Qwen3-Coder-480B and Claude 4 Sonnet. In particular, KAT-Coder attains 73.4 on SWE-Bench Verified when evaluated under Claude Code.

These results validate the effectiveness of our four-stage training pipeline (Mid-Term Training → SFT → RFT → RL). Each phase contributes distinct improvements—Mid-Term broadens reasoning depth, SFT enhances cross-language generalization, RFT stabilizes policy learning through relative rewards, and RL with Trie-Packed Training yields high-efficiency multi-trajectory optimization. The overall performance indicates that KAT-Coder not only matches proprietary systems on general reasoning and coding tasks but also demonstrates superior adaptability in authentic agentic coding scenarios.

\section{Conclusion}
In this report, we have detailed the design, training, and adaptation pipeline of KAT-Coder, a large-scale agentic code model developed to bridge the divide between research-grade models and deployable coding agents.

Through a structured four-stage process—Mid-Term Training, SFT, RFT, and Reinforcement-to-Deployment Adaptation—KAT-Coder achieves consistent improvements in reasoning stability, tool reliability, and contextual understanding. The integration of real software artifacts, synthetic reasoning traces, and production-grade agentic trajectories allows the model to operate effectively in complex, real-world development workflows.
Our findings demonstrate that agentic capability does not emerge from a single phase of optimization, but rather through the cumulative interaction between data diversity, reasoning supervision, and reinforcement alignment. The success of KAT-Coder highlights the importance of mid-term reasoning enrichment, structured multi-dimensional datasets, and robust RL adaptation in achieving deployable intelligence.

Looking ahead, future work will explore multi-modal agentic collaboration (e.g., code execution, GUI manipulation, and document editing), long-horizon memory persistence, and hierarchical planning architectures that allow models like KAT-Coder to serve as fully autonomous software collaborators in complex engineering environments.

\newpage
\bibliography{main}


\end{document}